\documentclass[11pt]{article}

\usepackage[final]{acl}

\usepackage{times}
\usepackage{latexsym}
\usepackage{booktabs}
\usepackage[T1]{fontenc}

\usepackage[utf8]{inputenc}

\usepackage{microtype}
\usepackage{amsmath}
\usepackage{amssymb}
\usepackage{inconsolata}
\usepackage{multirow}
\usepackage{enumitem}
\usepackage[super]{nth}

\usepackage{graphicx}

\usepackage{makecell}

\usepackage{fontawesome}

\usepackage{tikz}
\usetikzlibrary{positioning, fit, shapes.geometric, arrows.meta}
\title{TTLab at AlexandriaX-2026: A Fine-Tuned Surface Tagger for Arabic Machine-Translation Error-Span Detection and Classification}

\author{Ali Abusaleh, Bhuvanesh Verma, Alexander Mehler \\
Text Technology Lab (TTLab),  \\
         Goethe University Frankfurt\\
         \{a.abusaleh,verma,mehler\}@em.uni-frankfurt.de
         }

\begin{document}
\maketitle

\begin{abstract}
We present TTLab's submission to the AlexandriaX-2026 Subtask~3 on Arabic MT error span detection and classification. 
Our system frames the task as token-level classification over surface forms, preserving character offsets to ensure exact alignment with the evaluation metric. 
To handle severe label imbalance, we employ a focal loss with class weighting and dialect-specific decoding thresholds. Among six Arabic pre-trained encoders, MARBERTv2 achieves the best overall performance of 40.8 and 40.91 on the development and test set, respectively, ranking \nth{3} out of all participating teams. 
While our system localizes error spans effectively, classification of rare error types remains challenging, highlighting the need for data augmentation for tail categories. The code is available at {\href{https://github.com/ENTAILab/arabic-dialectal-mt-error-span-detection}{\faGithub~
TTLab at AlexandriaX-2026}}.
\end{abstract}

\section{Introduction}
\label{sec:intro}
Machine translation (MT) into dialectal Arabic remains substantially harder than into Modern Standard Arabic (MSA)~\cite{alabdullah:et:al:2025:advancingdialectalarabicmodern}, and sentence-level quality estimation is often too coarse for downstream applications like post-editing. Dialectal Arabic poses unique challenges: non-standard orthography, dialect-specific morphosyntactic patterns, and scarce parallel corpora. Fine-grained error span detection (locating exact erroneous substrings and labeling their category) provides the granularity these workflows require~\cite{math11194169}, but remains under-explored for Arabic dialects.
AlexandriaX-2026 Subtask~3{~\cite{alexandria2026}} addresses this gap: given an English source and its Arabic dialect MT output, systems must return every inaccurate character span with a category from a six-type linguistically motivated inventory~\citep{magdy-etal-2026-lqm}. The task spans five English-to-dialect directions and is evaluated at the exact character-offset level.
The \textbf{official baseline fine-tunes a 3B-parameter generative LLM} (NileChat) to produce JSON-encoded spans. However, this generative paradigm introduces three inefficiencies: (i) brittle JSON parsing (malformed output yields empty predictions), (ii) unreliable offset prediction (models must learn to generate character indices as text), and (iii) poor sample efficiency (completion-only loss ignores the discriminative signal from 80\% non-error tokens). Our surface-tagging formulation directly addresses these by reframing the task as discriminative token classification.
We present \textbf{TTLab's submission: a surface-tagging approach} that (i) preserves character offsets via sub-word tokenization, (ii) uses focal loss with down-weighted no-error labels to handle severe imbalance, and (iii) employs dialect-specific confidence thresholds. Among six Arabic pre-trained encoders, MARBERTv2 achieves the best performance (40.8 on dev, 40.9 on test), substantially outperforming its peers.

The remainder of this paper is organized as follows. Section~\ref{sec:related} reviews relevant work on MT quality estimation and Arabic dialectal NLP. Section~\ref{sec:data} describes the task and dataset. Section~\ref{sec:method} presents our system architecture and training methodology. Section~\ref{sec:experiments} details our experimental setup, and Section~\ref{sec:results} reports and analyzes the results. Finally, Section~\ref{sec:conclusion} concludes and discusses future directions.

\section{Related Work}
\label{sec:related}

\paragraph{MT error annotation with MQM/LQM.} The Multidimensional Quality Metrics (MQM) framework defines an open, extensible vocabulary of translation error types applicable to human and machine translation alike~\cite{mqm-lommel}. 
\citet{freitag-etal-2021-experts} established MQM-grounded expert error annotation as the reference methodology for MT evaluation, re-scoring top WMT systems with professional annotators. 
The LQM-style category scheme of AlexandriaX-2026 Subtask~3~\cite{alexandria2026} follows this tradition: errors are annotated as text spans, each carrying a category label.
\paragraph{Fine-grained quality estimation and error-span detection.} Predicting such annotations automatically has moved from sentence-level scores to fine-grained outputs. 
The WMT~2022 QE shared task adopted MQM annotations for sentence- \emph{and} word-level quality prediction~\cite{zerva-etal-2022-findings}, and WMT~2023 introduced a dedicated fine-grained \emph{error span detection} task, asking systems to predict error spans rather than binary \textsc{ok/bad} tags~\cite{wmt23-qe-task2}.
Learned metrics have followed: xCOMET couples sentence-level evaluation with the detection \emph{and categorisation} of error spans~\cite{guerreiro-etal-2024-xcomet}, while GEMBA-MQM prompts GPT-4 to mark MQM error spans without references~\cite{kocmi-federmann-2023-gemba}.
Our task shares this span-plus-category formulation, but targets \emph{dialectal Arabic} as an under-resourced target language with only $\sim$1.1k training sentences.
\paragraph{Sequence tagging with Arabic encoders.} Casting error identification as token-level tagging rather than generation has strong precedent: MaTESe reframes MT evaluation as a sequence-tagging problem over error spans~\cite{perrella-etal-2022-matese}, and GECToR showed that a tag-based encoder outperforms rewriting for error-focused tasks~\cite{omelianchuk-etal-2020-gector}. 
Our surface tagger follows this line, a token classifier over the raw
MT output whose offsets recover exact character spans, with a focal loss~\cite{lin:et:al:2018:focalloss} against the extreme \textsc{o}-class imbalance, built on Arabic pre-trained encoders: MARBERT/MARBERTv2, pre-trained on diverse Arabic varieties including dialectal text~\cite{abdul-mageed-etal-2021-arbert}; CAMeLBERT, with variant-specific models for MSA, dialectal, and classical Arabic~\cite{inoue-etal-2021-interplay}; and AraBERT, which established the value of Arabic-specific pre-training over multilingual models~\cite{antoun2020arabert}. Consistent with its dialectal pre-training, we find the MARBERTv2 family strongest for tagging dialectal MT output
(Section~\ref{sec:results}).

\section{Task and Data}
\label{sec:data}
We use the official dataset{~\cite{alexandria2026}}. Each instance consists of an English source sentence, a machine-generated Arabic dialect translation (the \textit{model\_prediction}), and a list of gold error annotations.
Every error is specified by its verbatim text span, character-level start and end offsets into the translation, and a category from the set \{graphetics, morphosyntax, orthography, pragmatics, semantics, sociolinguistics\} based on \citealt{magdy-etal-2026-lqm}. The data are organized by dialect direction: \texttt{ENG\_EGY}, \texttt{ENG\_MAU}, \texttt{ENG\_MOR}, \texttt{ENG\_PAL}, and \texttt{ENG\_UAE}.

Table~\ref{tab:dataset-stats} summarizes the training set statistics. Two properties dominate every design decision.
First, the data are \emph{small}: 1{,}125 training sentences contain only 1{,}997 error spans (mean 1.78 per sentence).
Second, the label distribution is heavily imbalanced: roughly 80\% of all tokens carry no error, and the category distribution ranges from \textit{sociolinguistics} (57.9\%) down to \textit{graphetics} (only 2 spans in the entire training set).
We therefore collapse the three rarest categories \{\textit{orthography}, \textit{pragmatics}, and \textit{graphetics}\} into a single \textit{other} class during training, while retaining \textit{morphosyntax}, \textit{semantics}, and \textit{sociolinguistics} as explicit categories. The collapsed \textit{other} class is expanded back to the most frequent original category at post-processing for scoring.
\begin{table}[t]

\centering
\small
\begin{tabular}{llr}
\toprule
\textbf{Split / property} & & \textbf{Value} \\
\midrule
\multirow{4}{*}{Sizes}
  & Train sentences & 1{,}125 \\
  & Dev sentences & 138 \\
  & Train error spans & 1{,}997 \\
  & Errors / sentence (mean) & 1.78 \\
\midrule
\multirow{5}{*}{Directions}
  & \texttt{ENG\_EGY} & 263 \\
  & \texttt{ENG\_MAU} & 244 \\
  & \texttt{ENG\_PAL} & 218 \\
  & \texttt{ENG\_UAE} & 231 \\
  & \texttt{ENG\_MOR} & 169 \\
\midrule
\multirow{6}{*}{Categories}
  & sociolinguistics & 57.9\% \\
  & semantics & 24.3\% \\
  & morphosyntax & 8.2\% \\
  & orthography & 5.3\% \\
  & pragmatics & 4.2\% \\
  & graphetics & 0.1\% \\
\bottomrule
\end{tabular}
\caption{AlexandriaX Subtask~3 training set statistics. Percentages are computed over all error spans.}
\label{tab:dataset-stats}
\end{table}

\begin{figure}[t]
  \centering
  \resizebox{\linewidth}{!}{%
  \begin{tikzpicture}[
      font=\small,
      node distance=0.45cm and 0.5cm,
      block/.style={draw, rectangle, rounded corners, fill=blue!5, minimum height=0.5cm, align=center},
      encoder/.style={draw, rectangle, fill=gray!10, minimum height=0.7cm, inner sep=0pt},
      token/.style={draw, rectangle, fill=white, minimum height=0.6cm, minimum width=1.0cm},
      tag/.style={draw, rectangle, rounded corners, minimum height=0.6cm, minimum width=1.0cm, fill=green!6},
      arrow/.style={-Stealth, thick}
  ]
      \node[token] (w1) {$w_1$};
      \node[token, right=of w1] (w2) {$w_2$};
      \node[token, right=of w2] (w3) {$w_3$};
      \node[token, right=of w3] (w4) {$w_4$};
      \node[token, right=of w4] (w5) {$w_n$};
      \node[fit=(w1)(w5), inner sep=2pt,
            label={[font=\bfseries]below:{Raw Arabic MT output $m$ (sub-word tokens, with char offsets)}}] (inp){};

      \node[encoder, fit={([yshift=0.8cm]w1.north west)([yshift=1.55cm]w5.north east)},
            label={[font=\small\bfseries]center:{Fine-tuned Arabic Encoder (e.g.\ MARBERTv2)}}] (enc){};

      \foreach \i in {1,...,5}{
          \node[token, fill=gray!4, anchor=south] (h\i) at ([yshift=0.8cm]enc.north -| w\i) {$\mathbf{h}_{\i}$};
      }

      \foreach \i in {1,...,5}{
          \node[block, above=0.65cm of h\i, minimum width=1.0cm] (c\i) {$W\mathbf{h}_{\i}$};
      }

      \node[tag, above=0.65cm of c1] (o1){O};
      \node[tag, above=0.65cm of c2, fill=red!10] (o2){B\textsc{-sem}};
      \node[tag, above=0.65cm of c3, fill=red!10] (o3){I\textsc{-sem}};
      \node[tag, above=0.65cm of c4] (o4){O};
      \node[tag, above=0.65cm of c5] (o5){O};
      \node[fit=(o1)(o5), inner sep=2pt,
            label={[font=\bfseries]above:{Per-token labels $\to$ merged into character spans $(s,e,c)$}}] (outp){};

      \node[block, fill=orange!8, right=1.3cm of o5, align=left] (thr){Decode:\\ per-direction\\ threshold $\tau_d$\\ + post-proc};

      \foreach \i in {1,...,5}{ \draw[arrow] (w\i.north) -- (w\i.north |- enc.south); }
      \foreach \i in {1,...,5}{
          \draw[arrow] (enc.north -| h\i.south) -- (h\i.south);
          \draw[arrow] (h\i) -- (c\i);
          \draw[arrow] (c\i) -- (o\i.south);
      }
      \draw[arrow, dashed, draw=orange!70!black] (thr.west) -- (o5.east);
  \end{tikzpicture}
  }
  
  \caption{Architecture of the error span detection and classification system.
           The raw Arabic MT output is tokenized with character offsets (bottom row) and encoded by a fine-tuned Arabic encoder.
           A linear head classifies each sub-word token; contiguous non-\textsc{O} tokens of the same category are merged and mapped to character spans via the stored offsets.
           The decoding block applies dialect-specific confidence thresholds and light post-processing.}
  \label{fig:architecture}
\end{figure}
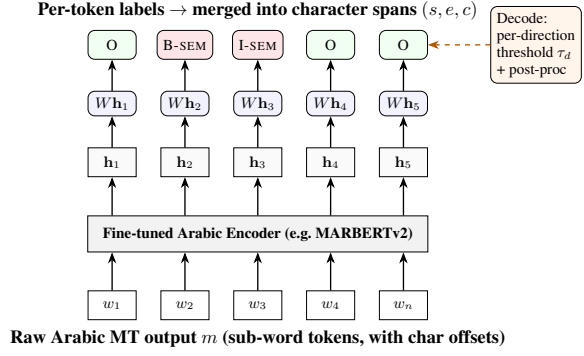

\section{System Overview}
\label{sec:method}

\subsection*{Tokenization and Surface Representation}
The core design choice of our system is to tag the \emph{surface} form of the MT output exactly as produced by the encoder's sub-word tokenizer, preserving the character offsets that map each token back to the original string.
We deliberately do not apply any normalization, stemming, or diacritic removal, because the evaluation compares predicted and gold spans by their character offsets in the original translation.
Any transformation that shifts character positions would break the alignment between a predicted token and the exact span expected by the metric. Tagging the raw surface thus keeps the offset bridge exact.

Figure~\ref{fig:architecture} illustrates the full pipeline. The raw Arabic MT output \(m\) is tokenized into sub-word units \(w_1,\dots,w_n\) together with their character offsets \((s_i, e_i)\).
These tokens are fed to a fine-tuned Arabic encoder, which produces contextual hidden states \(\mathbf{h}_i = f(m)_i \in \mathbb{R}^d\). A linear classification head is applied independently to each token, yielding label scores \(\mathbf{z}_i = W\mathbf{h}_i\) with \(W \in \mathbb{R}^{|\mathcal{L}| \times d}\).
The label set is \(\mathcal{L} = \{\text{O}\} \cup \mathcal{C}\), where \(\mathcal{C}\) is the collapsed set of error categories.

\subsection*{Joint Error Detection and Classification}
We formulate the task as per-token classification. Each gold error span induces the corresponding label for every token it overlaps; all other tokens receive the \textsc{O} (no error) label.
During inference, contiguous non-\textsc{O} tokens of the same category are merged into a single error span, and the stored character offsets are used to map it back to the exact character positions \((s, e, c)\) required by the scorer.

\subsection*{Imbalance-Aware Objective}
Because the \textsc{O} label dominates, a standard cross-entropy loss would cause the model to over-predict "no error." We therefore minimize a focal loss~\cite{lin:et:al:2018:focalloss} with down-weighted \textsc{O}:
\begin{equation*}
  \mathcal{L} = -\frac{1}{N}\sum_{i} \alpha_{y_i} (1 - p_{i,y_i})^{\gamma} \log p_{i,y_i},
\end{equation*}
where \(p_{i} = \mathrm{softmax}(\mathbf{z}_i)\) and \(y_i\) is the gold label. We set the focusing parameter \(\gamma = 2\) and use class weights \(\alpha_{\text{O}} = 0.3\), \(\alpha_{c} = 1\) for \(c \in \mathcal{C}\); padding and special tokens are excluded from the loss computation.
The down-weighting of \textsc{O} prevents the 80\% non-error majority from dominating the gradient, while the focal term sharpens the focus on hard, rare error tokens.

\subsection*{Decoding and Dialect-Specific Thresholds}
At inference time we compute the per-token error probability \(1 - p_{i,\text{O}}\). A token is flagged as an error when this probability exceeds a dialect-specific threshold \(\tau_d\), and its category is determined by \(\arg\max_{c\in\mathcal{C}} p_{i,c}\).
Contiguous tokens with the same category are merged, and their offsets define the character span. The thresholds \(\{\tau_d\}\) are tuned on out-of-fold (OOF) cross-validation predictions to maximize each dialect’s overall score.
Using a separate \(\tau_d\) per direction is critical because the dialects exhibit substantially different error densities (see Table~\ref{tab:dataset-stats}); a single global threshold would systematically over- or under-predict in different dialects.
A light post-processing step trims leading/trailing whitespace and punctuation from the predicted spans and discards spans shorter than two characters. For final scoring, the collapsed \textit{other} class is mapped to the most frequent original category among the merged tokens.

\section{Experiments}
\label{sec:experiments}

\paragraph{Setup.}
We fine-tune each encoder end-to-end using AdamW with a learning rate of \(2\times 10^{-5}\), batch size 16, and maximum sequence length 192. Gradient clipping is set to 1.0, and all models are trained with the focal loss defined in Section~\ref{sec:method}.
Results are reported as five-fold cross-validation on the training set (out-of-fold, OOF). The official development set (138 sentences) is used only as a secondary reference for backbone selection and diagnostics because single-split scores at this size are noisy; the one system decision made on dev is the character-level ensemble's vote threshold $v$ (see the ``Character-Level Voting Ensemble'' paragraph below), which has no OOF analogue in our pipeline.

\paragraph{Backbone Comparison.}
We compare six Arabic pre-trained encoders within the identical surface-tagging pipeline:
\textsc{MARBERTv2}~\cite{abdul-mageed-etal-2021-arbert},
a \textsc{MARBERTv2} cross-encoder checkpoint fine-tuned on QuranQA\footnote{\url{https://huggingface.co/yoriis/marbertv2-crossencoder-quranqa25}},
\textsc{CAMeLBERT-DA}~\cite{inoue-etal-2021-interplay},
\textsc{SaudiBERT}~\cite{qarah2024saudibert},
\textsc{AraBERTv2}~\cite{antoun2020arabert},
and \textsc{NileChat-3B}~\cite{el-mekki-etal-2025-nilechat} used as a token tagger.
This sweep selects the final backbone; Table~\ref{tab:main} reports the results.

\paragraph{Per-Dialect Thresholds and Post-Processing.}
Tuning a separate confidence threshold for each dialect, rather than a single global threshold, yields consistent gains of up to 0.8 points in overall OOF score.
The light post-processing (span trimming and minimum length filtering) adds a further 0.3-0.5 points.
Both components are retained in the final system.

\paragraph{Character-Level Voting Ensemble.}
As an extension, we build a character-level voting ensemble on top of the six per-backbone taggers. Each model independently decodes its spans; a character position is considered part of an error if at least \(v\) models agree, and the category is decided by a weighted majority vote.
Sweeping the vote threshold \(v\) interpolates between the union (higher recall) and intersection (higher precision) of the individual taggers. The best operating point on the dev set is reported in Table~\ref{tab:main}.

\section{Results}
\label{sec:results}

\begin{table}[t]
\centering
\small
  \resizebox{\linewidth}{!}{%
  \begin{tabular}{llcc}
  \toprule
  \textbf{System} & \textbf{Backbone} & \textbf{OOF} & \textbf{Dev} \\
  \midrule
  \multicolumn{4}{l}{\textit{Fine-tuned surface tagger}} \\
  Surface tagger & NileChat-3B            & 28.2 & 29.2 \\
  Surface tagger & AraBERTv2              & 34.8 & 34.4 \\
  Surface tagger & SaudiBERT              & 36.7 & 35.1 \\
  Surface tagger & CAMeLBERT-DA           & 36.7 & 35.8 \\
  Surface tagger & \makecell{MARBERTv2  (QuranQA c.e.)} & 38.9 & 38.5 \\
  \textbf{Surface tagger (final)} & \textbf{MARBERTv2} & \textbf{39.4} & \textbf{40.8} \\
  \midrule
  \multicolumn{4}{l}{\textit{Extension built on top}} \\
  Char-vote ensemble & six backbones & -   & 39.9 \\
  \bottomrule
\end{tabular}}
\caption{Results on AlexandriaX Subtask~3. We report
         \((\text{Overlap-span }F_1 + \text{Error-Category micro-}F_1)/2\) (in percent); the official
         leaderboard instead combines overlap-span \(F_1\) and error-class \(F_1\) as
         \((\text{Overlap}+\text{Class})/2\), each macro-averaged across the five dialect directions
         (Exact-Match \(F_1\) is reported by the leaderboard separately as a diagnostic and does not
         enter the ranking score), so our development scores here are a pooled/micro proxy and are not
         directly comparable.
         OOF is five-fold cross-validation on the training set; Dev is the official 138-sentence development set.
         \textsc{MARBERTv2} achieves the best performance on both splits and is our submitted system.}
\label{tab:main}
\end{table}

\paragraph{Backbone Comparison.}
Table~\ref{tab:main} shows that \textsc{MARBERTv2} is the strongest backbone, achieving 39.4 OOF and 40.8 Dev.
It is followed by the \textsc{MARBERTv2} cross-encoder checkpoint (38.9/38.5), and then by the dialectal \textsc{CAMeLBERT-DA} and \textsc{SaudiBERT} (both 36.7 OOF). 
\textsc{AraBERTv2} and the generative \textsc{NileChat-3B} perform considerably worse. The \textsc{MARBERTv2} family clearly dominates, and we select it as the final backbone.

\paragraph{Localization Outperforms Categorization.}
Decomposing the metric reveals a consistent pattern: at the best operating points of both single models and the ensemble, the overlap-span \(F_1\) reaches 46-48, while the error-category micro-\(F_1\) is only 31-33.
A matched-span analysis indicates that when the model correctly localizes an error, it usually assigns the correct category.
The residual category loss is driven primarily by \emph{undetected} spans (which are category false negatives) and by the extremely rare classes (2-83 training examples each), not by misclassification of detected spans.

\paragraph{Threshold-Tuning Validity: Nested vs.\ Pooled OOF.}
Because \(\{\tau_d\}\) in Section~\ref{sec:method} is tuned and evaluated on the same pooled OOF predictions, we additionally run a nested (leave-one-fold-out) variant: for each fold, thresholds are tuned only on the other four folds' OOF predictions before scoring the held-out fold. For the submitted \textsc{MARBERTv2} configuration this lowers the OOF overall score from 39.4 (pooled, as in Table~\ref{tab:main}) to \textbf{35.4} (nested), a \(\sim\)4-point gap that we attribute to the pooled procedure's mild overfitting to the OOF set it is also scored on. We take 35.4 to be the more honest OOF estimate and report it alongside the original in Table~\ref{tab:ablation}.

\paragraph{Loss and Category-Granularity Ablations.}
Table~\ref{tab:ablation} isolates two design choices. Replacing focal loss with weighted cross-entropy loses 0.8 nested-OOF and 1.9 dev-macro-\(F_1\) points, confirming focal weighting is a modest but consistent contributor. More notably, training on all six categories directly (no collapse) matches or exceeds the submitted configuration on every metric, including dev macro-\(F_1\) (18.2 vs.\ 16.3), because \textit{orthography} recovers a category-specific signal (OOF \(F_1\) 9.6 vs.\ 4.9 from the fixed fallback) and \textit{pragmatics} moves off zero; \textit{graphetics} stays at 0 (2 training examples). We confirmed this on the official leaderboard by retraining and resubmitting: the no-collapse system raises the test Overall Score from \textbf{40.91} (submitted) to \textbf{42.00} (Table~\ref{tab:perdir})
\begin{table}[t]
\centering
\small
\begin{tabular}{lccc}
\toprule
\textbf{Direction} & \textbf{Exact} & \textbf{Overlap} & \textbf{Class} \\
\midrule
ENG\_EGY & 14 & 50 & 47 \\
ENG\_MOR & 17 & 47 & 32 \\
ENG\_MAU &  7 & 39 & 44 \\
ENG\_PAL & 16 & 47 & 28 \\
ENG\_UAE & 10 & 49 & 42 \\
\midrule
\textbf{Macro} & \textbf{13} & \textbf{46} & \textbf{38} \\
\bottomrule
\end{tabular}
\caption{Official Codabench test-set scores (\%) per dialect direction for the revised focal, no-collapse \textsc{MARBERTv2} system. Overall Score $(\text{Overlap}_{\text{macro}} + \text{Class}_{\text{macro}})/2 = \mathbf{42.00}$. Exact-Match is a diagnostic and does not enter the ranking score.}
\label{tab:perdir}
\end{table}

\begin{table}[t]
\centering
\small
\resizebox{\linewidth}{!}{%
\begin{tabular}{lccccc}
\toprule
\textbf{Config} & \textbf{OOF$_{\text{pooled}}$} & \textbf{OOF$_{\text{nested}}$} & \textbf{Dev} & \textbf{Dev macro-}$F_1^{\text{cat}}$ \\
\midrule
Focal + collapse (submitted) & 39.4 & 35.4 & 38.9 & 16.3 \\
Cross-entropy + collapse     & 39.1 & 34.6 & 38.8 & 14.5 \\
Focal + no collapse          & 39.3 & \textbf{35.9} & \textbf{40.1} & \textbf{18.2} \\
\bottomrule
\end{tabular}}
\caption{Ablations on the final \textsc{MARBERTv2} tagger (5-fold OOF, retrained per configuration). OOF$_{\text{pooled}}$ matches the procedure in Table~\ref{tab:main}; OOF$_{\text{nested}}$ tunes per-dialect thresholds on the other four folds only. Dev macro-$F_1^{\text{cat}}$ is the true per-category macro-$F_1$ (over all six original categories) on the official dev set, using overlap-based span matching.}
\label{tab:ablation}
\end{table}

\paragraph{Per-Category Breakdown.}
Table~\ref{tab:percat} reports precision, recall, and \(F_1\) for each of the six original categories, computed on the 5-fold OOF predictions of the submitted (focal, collapsed) configuration over the \emph{training} set. We use the training set rather than dev for this analysis because the 138-sentence dev set happens to contain zero gold spans of \textit{orthography}, \textit{pragmatics}, or \textit{graphetics}, making it unusable for assessing exactly the rare categories this analysis is meant to cover. \textit{Sociolinguistics} and \textit{semantics}, the two head categories, reach \(F_1\) 39.2 and 23.6 respectively. \textit{Morphosyntax} is modeled explicitly (it is \emph{not} collapsed) yet still only reaches \(F_1\) 5.3, a genuine weak point we attribute to its smaller size (164 spans) and higher surface-form variability rather than to the collapse strategy. \textit{Orthography}, which \emph{is} collapsed into \textit{other} at train time, inherits \(F_1\) 4.9 purely via the fixed fallback label; \textit{pragmatics} and \textit{graphetics} are never recovered (\(F_1\)=0).

\begin{table}[t]
\centering
\small
\begin{tabular}{lrrrr}
\toprule
\textbf{Category} & \textbf{Supp.} & \textbf{P} & \textbf{R} & \(F_1\) \\
\midrule
sociolinguistics & 1157 & 31.7 & 51.3 & 39.2 \\
semantics        & 485  & 21.0 & 27.0 & 23.6 \\
morphosyntax     & 164  & 5.8  & 4.9  & 5.3 \\
orthography      & 106  & 5.2  & 4.7  & 4.9 \\
pragmatics       & 83   & 0.0  & 0.0  & 0.0 \\
graphetics       & 2    & 0.0  & 0.0  & 0.0 \\
\bottomrule
\end{tabular}
\caption{Per-category precision/recall/$F_1$ (\%) of the submitted (focal, collapsed) \textsc{MARBERTv2} tagger, on 5-fold OOF predictions over the \emph{training} set. Support is the number of gold spans of that category. \textit{Orthography} is collapsed into \textit{other} at train time, so its score here reflects the fixed \textit{other}$\to$\textit{orthography} fallback rather than category-specific discrimination; \textit{morphosyntax} is modeled explicitly and its low score is a genuine gap (see Table~\ref{tab:ablation} for the un-collapsed alternative).}
\label{tab:percat}
\end{table}

\paragraph{The Ensemble Does Not Improve Over the Single Best Model.}
The character-level voting ensemble attains a dev score of 39.9 at its optimal vote threshold, improving recall and category \(F_1\) slightly but failing to exceed the single \textsc{MARBERTv2} tagger (40.8).
Because \textsc{MARBERTv2} dominates its peers, blending in weaker, correlated encoders only dilutes its predictions.
We therefore submit the single \textsc{MARBERTv2} surface tagger as our final system.

\section{Conclusion}
\label{sec:conclusion}
We presented a surface-tagging approach for dialectal Arabic MT error span detection and classification that operates directly on raw, unnormalized text.
By preserving character offsets through the tokenization and modeling pipeline, we ensure exact alignment with the gold spans required by the evaluation metric. A focal loss with class weighting effectively handles the severe label imbalance, and dialect-specific confidence thresholds adapt to varying error densities.
Among six pre-trained Arabic encoders, \textsc{MARBERTv2} yields the best results, and a character-level ensemble offers no further gain. Detailed analysis reveals that the system localizes errors well but struggles with the rarest categories due to extreme data scarcity.
Future work could explore data augmentation for tail categories, cross-lingual transfer from higher-resource languages, and exact-match-span optimization to better align with the official scoring regime.

\clearpage
\newpage

\section*{Limitations}
\label{sec:limitations}
Several limitations of this work should be noted.
First, the training set is extremely small (1,125 sentences) and contains only two examples of the \textit{graphetics} category, making it nearly impossible to learn that class reliably. Our collapse strategy mitigates this but does not fully solve the data sparsity problem.
Second, the system relies on a single, relatively large encoder (\textsc{MARBERTv2}) and may not scale gracefully to resource-constrained environments.
Third, although we deliberately avoid normalization to preserve character offsets, this choice ties the system to the exact tokenization of the chosen encoder; a change of tokenizer would require re-mapping the offset annotations.
Finally, our development metric pools overlap-span \(F_1\) and micro-averaged category \(F_1\) across the whole split, whereas the confirmed official metric computes overlap-span \(F_1\) and error-class \(F_1\) \emph{per dialect direction} and macro-averages each across the five directions before combining them; exact-match \(F_1\) (literal character-offset equality) is reported by the leaderboard separately as a diagnostic that does not enter the ranking score. Our development-set numbers are therefore a pooled/micro proxy and are not directly comparable to the official leaderboard; we did not optimize for the official aggregation directly.
\section*{Acknowledgments}
This research is partially funded by the German Research Foundation 
within the Infrastructure Priority Programme \emph{New Data Spaces for the Social Sciences} \href{https://www.dfg.de/de/aktuelles/neuigkeiten-themen/info-wissenschaft/2023/info-wissenschaft-23-20}{(SPP 2431)}, Research-driven Infrastructure for Advanced Survey-related Data (CIRCLET) measure project number ~\href{https://gepris.dfg.de/project/539634240}{539634240} and (Semi-)Automated thematic text classification as a basis for corpus-linguistic value-added services (Project number: \href{https://gepris.dfg.de/project/531750631}{531750631}).

\bibliography{custom}


\end{document}